# BlitzNet: A Real-Time Deep Network for Scene Understanding

Nikita Dvornik*  Konstantin Shmelkov*  Julien Mairal  Cordelia Schmid

Inria[†]

## Abstract

*Real-time scene understanding has become crucial in many applications such as autonomous driving. In this paper, we propose a deep architecture, called BlitzNet, that jointly performs object detection and semantic segmentation in one forward pass, allowing real-time computations. Besides the computational gain of having a single network to perform several tasks, we show that object detection and semantic segmentation benefit from each other in terms of accuracy. Experimental results for VOC and COCO datasets show state-of-the-art performance for object detection and segmentation among real time systems.*

## 1. Introduction

Object detection and semantic segmentation are two fundamental problems for scene understanding in computer vision. The task of object detection is to identify on an image all objects of predefined categories and localize them via bounding boxes. Semantic segmentation operates at a finer scale; its aim is to parse an image and associate a class label to each pixel. Despite the similarities of the two tasks, only few works have tackled them jointly [3, 11, 28, 29].

Yet, there is a strong motivation to address both problems simultaneously. On the one hand, good segmentation is sufficient to perform detection in some cases. As Figure 1 suggests, an object may be sometimes identified and localized from segmentation only by simply looking at connected components of pixels sharing the same label. In the more general case, it is easy to conduct a simple experiment showing that ground-truth segmentation is a meaningful clue for detection, using for instance ground-truth segmentation as the input of an object detection pipeline. On the other hand, correctly identified detections are also useful for segmentation as shown by the success of weakly supervised segmentation techniques that learn from bounding box annotation only [21]. The goal of our paper is to solve efficiently both problems at the same time, by exploiting image data annotated at the global object level (via bounding boxes), at the pixel level (via partially or fully annotated segmentation maps), or at both levels.

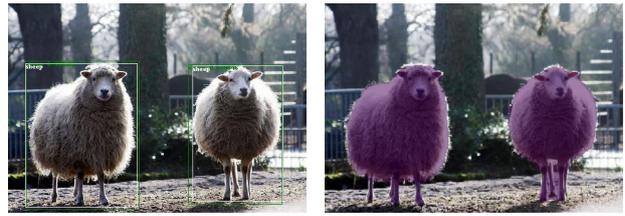

(a) Generated bounding boxes    (b) Generated masks

Fig. 1: The outputs of our pipeline. (a) The results of object detection. (b) The results of semantic segmentation.

As most recent image recognition pipelines, our approach is based on convolutional neural networks [12], which are widely adopted for object detection [6] and semantic segmentation [17]. More precisely, deep neural networks were first used as feature extractors to classify a large number of candidate bounding boxes [6], which is computationally expensive. The improved version [5] reduces the computational cost but relies on shallow techniques for extracting bounding box proposals and does not allow end-to-end training. This issue was later solved in [25] by making the object proposal mechanism a part of the neural network. Yet, the approach remains expensive and relies on a region-based strategy (see also [13]) that makes the network architecture inappropriate for semantic segmentation.

To match the real-time speed requirement, we choose instead to base our work on the Single Shot Detection (SSD) [16] approach, which consists of a fully-convolutional model to perform object detection in one forward pass. Besides the fact that it allows all computations to be performed in real time, the pipeline is more generic, imposes less constraints on the network architecture and opens new perspectives to solve our multi-task problem.

Interestingly, recent work on semantic segmentations are also moving in the same direction, see for instance [17]. Specific to semantic segmentation, [17] also introduces new ideas such as the joint use of feature maps of different resolutions, in order to obtain more accurate classification. The idea was then improved by adding deconvolutional layers at all scales to better aggregate context, in addition to using skip and residual connections [26]. Deconvolutional layers turned out to be useful to estimate precise segmentations, and are thus good candidates to design architectures where localization is important.

---

*The authors contributed equally.

[†]Univ. Grenoble Alpes, Inria, CNRS, Grenoble INP, LJK, 38000 Grenoble, France.

In this paper, we consider the multi-task scene understanding problem consisting of joint object detection and semantic segmentation. For that purpose, we propose a novel pipeline called BlitzNet, which will be released as an open-source software package. BlitzNet is able to provide accurate segmentation and object bounding boxes in real time. With a single network for solving both problems, the computational cost is reduced, and we show also that the two tasks benefit from each other in terms of accuracy.

The paper is organized as follows: Section 2 discusses related work; Section 3 presents our real-time multi-task pipeline called BlitzNet. Finally, Section 4 is devoted to our experiments, and Section 5 concludes the paper.

## 2. Related Work

Before we introduce our approach, we now present techniques for object detection, semantic segmentation, and previous attempts to combine both tasks.

**Object detection.** The field of object detection has been recently dominated by variants of the R-CNN architecture [5, 25], where bounding-box proposals are independently classified by a convolutional neural network, and then filtered by a non-maximum suppression algorithm. It provides great accuracy, but relatively low inference speed since it requires a significant amount of computation per proposal. R-FCN [13] is a fully-convolutional variant that further improves detection and significantly reduces the computational cost per proposal. Its region-based mechanism is however dedicated to object detection only.

SSD [16] is a recent state-of-the-art object detector, which uses a sliding window approach instead of generated proposals to classify all boxes directly. SSD creates a scale pyramid to find objects of various sizes in one forward pass. Because of its speed and high accuracy, we have chosen to build our work on, and subsequently improve, the SSD approach. Finally, YOLO [23, 24] also provides real-time object detection and shares some ideas with SSD.

**Semantic segmentation and deconvolutional layers.** Deconvolutional architectures consist of adding to a classical convolutional neural networks with feature pooling, a sequence of layers whose purpose is to increase the resolution of the output feature maps. This idea is natural in the context of semantic segmentation [20], since segmentation maps are expected to have the same resolution as input images. Yet, it was also successfully evaluated in other contexts, such as pose estimation [19], and object detection, as extensions of SSD [4] and Faster-R-CNN [14].

**Joint semantic segmentation and object detection.** The idea of joint semantic segmentation and object detection was investigated first for shallow approaches in [3, 18, 29, 7]. There, it was shown that learning both tasks simultaneously could be better than learning them independently.

More recently, UberNet [11] integrates multiple vision tasks such as semantic segmentation and object detection into a single deep neural network. The detection part is based on the Faster R-CNN approach and is thus neither fully-convolutional nor real-time. Closely related to our work, but dedicated to autonomous driving, [28] also proposes to integrate semantic segmentation and object detection via a deep network. There, the VGG16 network [27] is used to compute image features (encoding step), and then two different sub-networks are used for the prediction of object bounding boxes and segmentation maps (decoding).

Our work is inspired by these previous attempts, but goes a step further in integrating the two tasks, with a fully convolutional approach where network weights are shared for both tasks until the last layer, which has advantages in terms of speed, feature sharing, and simplicity for training.

## 3. Scene Understanding with BlitzNet

In this section, we introduce the BlitzNet architecture and discuss its different building blocks.

### 3.1. Global View of the Pipeline

The joint object detection and segmentation pipeline is presented in Figure 2. The input image is first processed by a convolutional neural network to produce a map that carries high-level features. Because of its high performance for classification and good trade-off for speed, we use the network ResNet-50 [9] as our feature encoder.

Then, the resolution of the feature map is iteratively reduced to perform a multi-scale search of bounding boxes, following the SSD approach [16]. Inspired by the hourglass architecture [19] for pose estimation and an earlier work on semantic segmentation [20], the feature maps are then up-scaled via deconvolutional layers in order to predict subsequently precise segmentation maps. Recent DSSD approach [4] uses a similar strategy for object detection the top part of our architecture presented in Figure 2 may be seen as a variant of DSSD with a simpler "deconvolution module", called ResSkip, that involves residual and skip connections.

Finally, prediction is achieved by single convolutional layers, one for detection, and one for segmentation, in one

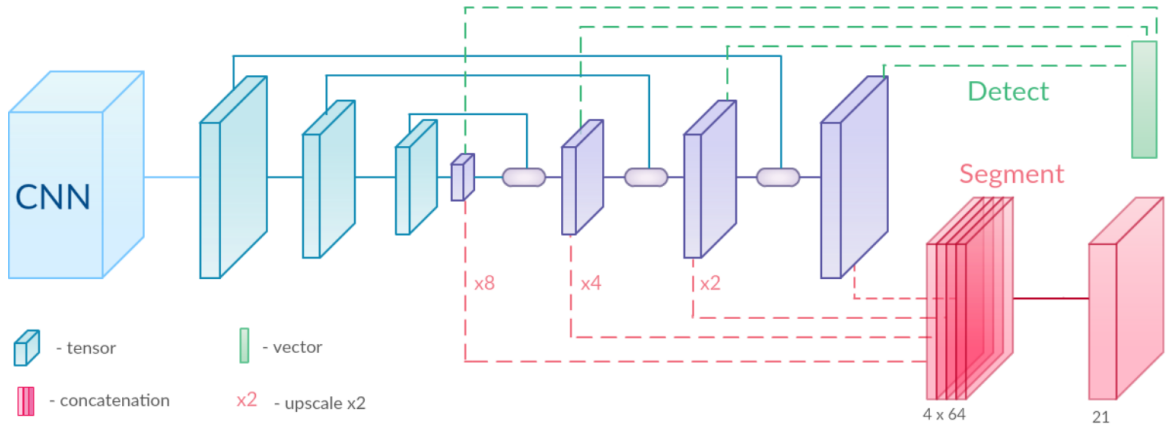

Fig. 2: The BlitzNet architecture, which performs object detection and segmentation with one fully convolutional network. On the left, CNN denotes a feature extractor, here ResNet-50 [9]; it is followed by the downscale-stream (in blue) and the last part of the net is the upscale-stream (in purple), which consists of a sequence of deconvolution layers interleaved with ResSkip blocks (see Figure 3). The localization and classification of bounding boxes (top) and pixelwise segmentation (bottom) are performed in a multiscale fashion by single convolutional layers operating on the output of deconvolution layers.

forward pass, which is the main originality of our work.

### 3.2. SSD and Downscale Stream

The Single Shot MultiBox Detector [16] tiles an input image with a regular grid of anchor boxes and then uses a convolutional neural network to classify these boxes and predict corrections to their initial coordinates. In the original paper [16], the base network VGG-16 [27] is followed by a cascade of convolutional and pooling layers to form a sequence of feature maps with progressively decreasing spatial resolution and increasing field of view. In [16], each of these layers is processed separately in order to classify and predict coordinates correction for a set of default bounding boxes of a particular scale. At test time, the set of predicted bounding boxes is filtered by non-maximum suppression (NMS) to form the final output.

Our pipeline uses such a cascade (see Figure 2), but the classification of bounding boxes and pixels to build the segmentation maps is performed in subsequent layers, called deconvolutional layers, which will be described next.

### 3.3. Deconvolution Layers and ResSkip Blocks

Modeling visual context is often a key to complicated scenes parsing, which is typically achieved by pooling layers in a convolutional neural network, leading to large receptive fields for each output neuron. For semantic segmentation, precise localization is equally important, and [20] proposes to use deconvolutional operations to solve that issue. Later, this process was improved in [19] by adding skip connections. Apart from combining high- and low-level features it also eases the learning process [9].

Like [4] for object detection and [19] for pose estimation, we also use such a mechanism with skip connections that combines feature maps from the downscale and upscale streams (see Figure 2). More precisely, maps from the downscale and upscale streams are combined with a simple strategy, which we call ResSkip, presented in Figure 3. First, incoming feature maps are upsampled to the size of corresponding skip connection via bilinear interpolation. Then both skip connection feature maps and upsampled maps are concatenated and passed through a block ($1 \times 1$ convolution, $3 \times 3$ convolution, $1 \times 1$ convolution) and summed with the upsampled input through a residual connection. The benefits of this topology will be justified and discussed in more details in the experimental section.

### 3.4. Multiscale Detection and Segmentation

The problem of semantic segmentation and object detection share several key properties. They both require per-region classification, based on the pixels inside an object while taking into account its surrounding, and benefit from rich features that include localization information. Instead of training a separate network to perform these two tasks, we train a single one that allows weight sharing, such that both tasks can benefit from each other.

In our pipeline, most of the weights are shared. Object detection is performed by a single convolutional layer that predicts a class and coordinate corrections for each bounding box in the feature maps of the upscale stream. Similarly, a single convolutional layer is used to predict the pixel la-bels and produce segmentation maps. To achieve this we upscale all the activations of the upscale stream, concatenate them and feed to the final classification layer.

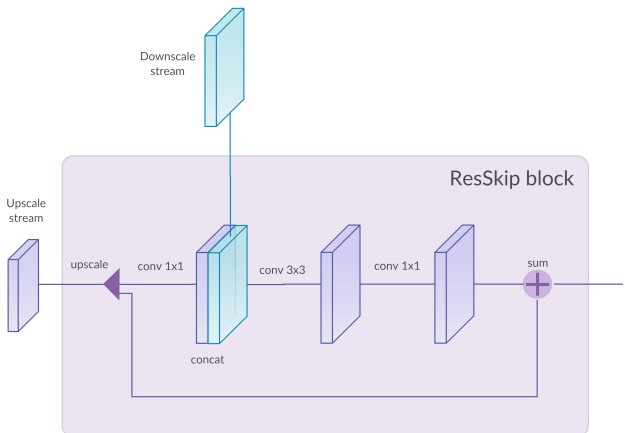

Fig. 3: ResSkip block integrating feature maps from the upscale and downscale streams, with skip connection.

### 3.5. Speeding up Non-Maximum Suppression

Increasing the number of anchor boxes heavily affects inference time because it performs NMS on a potentially huge number of proposals (in the worst case scenario, it may be all of them). Indeed, we observed that by using sliding window proposals, addition of small scale proposals slows down the inference even more than increasing image resolution. Surprisingly, non-maximum suppression may then become the bottleneck at inference time. We observed that this occurred sometimes for particular object classes that return a lot of bounding box candidates.

Therefore, we suggest a different post-processing strategy to accelerate detection when there are too many proposals. For each class, we pre-select the top 400 boxes with largest scores, and perform NMS leaving only 50 of them. Overall, the final detection is the top 200 highest scoring boxes per image after non-maximum suppression. This strategy yields a reasonable computational time for NMS, and has marginal impact on accuracy.

### 3.6. Training and Loss Functions

Given labeled training data where each data point is annotated with segmentation maps, or bounding boxes, or with both, we consider a loss function which is simply the sum of two loss functions of the two task. Note that we tried reweighting the two loss functions, but we did not observe noticeable improvements in terms of accuracy.

For segmentation, the loss is the cross-entropy between predicted and target class distribution of pixels [1]. Specifically, we use a $1 \times 1$ convolutional operation with 64 channels to map each layer of the upscale-stream to an intermediate representation. After this, each layer is upscaled to the size of the last layer using bilinear interpolation and all maps are concatenated. This representation is mapped to $c$ feature maps, where $c$ is the number of classes, by using $3 \times 3$ convolutions to predict posterior class probabilities.

For detection, we use the same loss function as [16] when performing tiling of the input image with anchor boxes and matching them to ground truth bounding boxes. We use activations of each layer in the upscale-stream to regress corrections for coordinates of the anchor boxes and to predict the class probability distribution. We use the same data augmentation suggested in the original SSD pipeline, namely photometric distortions, random crops, horizontal flips and zoom-out operation.

## 4. Experiments

We now present various experiments conducted on the COCO, Pascal VOC 2007 and 2012 datasets, for which both bounding box annotations and segmentation maps are available. Section 4.1 discusses in more details the datasets and the metrics we used; Section 4.2 presents technical details that are useful to make our work reproducible, and then each subsequent subsection is devoted to a particular experiment. The last two sections discuss the inference speed and clarify particular choices in the network architecture. Our code is now available as an open-source software package at http://thoth.inrialpes.fr/research/blitznet/.

### 4.1. Datasets and Metrics

We use the COCO [15], VOC07, and VOC12 datasets [2]. All images in the VOC datasets are annotated with ground truth bounding boxes of objects and only a subset of VOC12 is annotated with target segmentation masks. The VOC07 dataset is divided into 2 subsets, trainval (5011 images) and test (4952 images). The VOC12-train subset contains 5717 images annotated for detection and 1464 of them have segmentation ground truth as well (VOC12-train-seg), while VOC12-val has 5823 images for detection and 1449 images for segmentation (we call this subset VOC12-val-seg). Both datasets have 20 object classes.

The COCO dataset includes 80 object categories for detection and instance segmentation. For the task of detection, there are 80k images for training and 40k for validation. There is no either a protocol for evaluation of semantic segmentation or even annotations to train it from. In this work, we are interested particularly in semantic segmentation masks so we obtain them from instance segmentation annotations by combining instances of one category.

To carry out more extensive experiments we leverage extra annotations for VOC12 segmentation provided by [8], which gives a total of 10,582 fully annotated images for training that we call VOC12-train-seg-aug. We still keep the original PASCAL annotations in VOC12 val-seg, even if a more precise annotation is available in [8], for a fair comparison with other methods that do not benefit from these extra annotations.

In VOC12 and VOC07 datasets, a predicted bounding box is correct if its intersection over union with the ground truth bounding box is higher than 0.5. The metric for evaluation detection performance is the mean average precision (mAP) and the quality of predicted segmentation masks is measured with mean intersection over union (mIoU).

### 4.2. Experimental Setup

In this section, we discuss the common setup to all experiments. BlitzNet is coded in Python and TensorFlow. All experiments were conducted on a single Titan X GPU (Maxwell architecture), which makes the speed comparison with previous work easy, as long as they use the same GPU.

**Optimization Setup.** In all our experiments, unless explicitly stated otherwise, we use the Adam algorithm [10], with a mini-batch size of 32 images. The initial learning rate is set to $10^{-4}$ and decreased twice during training by a factor 10. We also use a weight decay parameter of $5 \times 10^{-4}$.

**Modeling setup.** As already mentioned, we use ResNet-50 [9] as a feature extractor, 512 feature maps for each layer in down-scale and up-scale streams, 64 channels for intermediate representations in the segmentation branches; BlitzNet300 takes input images of size $300 \times 300$ and BlitzNet512 uses $512 \times 512$ images. Different versions of the network vary in the stride of the last layer of the upscaling-stream. Strides 4 and 8 in the result tables are denoted as (s4) and (s8) suffix respectively.

### 4.3. PASCAL VOC 2007

In this experiment, we train our networks on the union of VOC07 trainval set and VOC12 trainval set; then, we test them on the VOC07 test set. The results are reported in the Table 1. For experiments that involve segmentation, we leverage ground truth segmentation masks during training if they are available in VOC12 train-seg-aug or in VOC12 val-seg. When using images of size $300 \times 300$ as input, the stochastic gradient descent algorithm is performed by training for 65K iterations with the initial learning rate, which is then decreased after 35K and 50K steps. When training on $512 \times 512$ images, we choose the batch size of 16 and learn for $75K$ iterations decreasing the learning rate after 45K and 60K steps.

The results show that BlitzNet300 outperforms SSD300 and YOLO with a 78.5 mAP, while being a real time detector. BlitzNet512 (s8) performs 0.8% better than R-FCN - the most accurate competitive model, scoring 81.2% mAP.

We further improve the results by training for detection and segmentation jointly achieving 79.1% and 81.5% mAP with BlitzNet300 (s4) and BlitzNet512 (s8) respectively.

We think that the performance gain for BlitzNet300 over BlitzNet512 could be explained by the larger stride used for the last layer, which is 4, vs 8 for BlitzNet512, and seems to be helpful for better learning finer details. Unfortunately, training BlitzNet512 with stride 4 was impossible because of memory limitations on our single GPU.

### 4.4. PASCAL VOC 2012

In this experiment, we use VOC12 train-seg-aug for training and VOC12 val-seg for testing both segmentation and detection. We train the models for 40K steps with the initial learning rate, and then decrease it after 25K and 35K iterations. As Table 3 shows, joint training improves accuracy on both tasks comparing to learning a single task. Detection improves by more than 1% while segmentation mIoU grows by 0.4%. We argue that this result could be explained by feature sharing in the universal architecture.

To confirm this fact, we conducted another experiment by adding the VOC07 trainval images to VOC12 train-seg-aug for training. Then, the proportion of images that have segmentation annotations to the ones that have detection ones only is 2/1, in contrast to the previous experiments where all the images where annotated for both tasks. To deal with cases where a mini-batch has no images to train for segmentation, we set the corresponding loss to 0 and do not back propagate with respect to these images, otherwise we use all images that have target segmentation masks in a batch to update the weights. The results presented in Table 4 show an improvement of 3.3%. Detection also improves in mAP by 0.6%. Figure 7 shows that extra data for detection helps to improve classification results and to mitigate confusion between similar categories. In Table 2, we report results for these models on the VOC12 test server, which again shows that our results are competitive. More qualitative results, including failure cases, are presented in the supplementary material.

### 4.5. Microsoft COCO Dataset

To further validate the proposed framework, we conduct experiments on the COCO dataset [15]. Here, as explained in Section 4.1, we obtain segmentation masks and again training the model on different types of data, i.e., detection, segmentation and both, to study the influence of joint training on detection accuracy.

We train the BlitzNet300 or BlitzNet512 models for 700k iterations in total, starting from the initial learning rate $10^{-4}$ and then decreasing it after the 400k and 550k iterations by the factor of 10. Table 5 shows clear benefits from joint training for both of the tasks on the COCO dataset. To be comparable with other methods, we also report the

| network | backbone | mAP | aero | bike | bird | boat | bottle | bus | car | cat | chair | cow | table | dog | horse | mbike | persn | plant | sheep | sofa | train | tv |
|---|---|---|---|---|---|---|---|---|---|---|---|---|---|---|---|---|---|---|---|---|---|---|
| SSD300* [16] | VGG-16 | 77.6 | 79.2 | 84.0 | 75.6 | 69.9 | 50.9 | 86.7 | 85.9 | 88.6 | 60.1 | 81.4 | 76.8 | 86.2 | 87.3 | 84.2 | 79.5 | 52.7 | 79.3 | 79.4 | 87.7 | 77.2 |
| SSD300* (our reimpl) | ResNet-50 | 75.3 | 75.3 | 85.1 | 72.5 | 67.4 | 45.5 | 85.7 | 83.9 | 82.8 | 57.2 | 79.1 | 76.7 | 83.1 | 86.5 | 83.3 | 77.5 | 50.1 | 74.4 | 79.4 | 86.5 | 73.3 |
| **BlitzNet300 (s8)** | ResNet-50 | 78.5 | 79.7 | 85.9 | 80.1 | 72.1 | 50.9 | 87.0 | 84.6 | 87.2 | 62.3 | 83.7 | 77.1 | 87.3 | 85.0 | 84.7 | 79.2 | 54.9 | 81.5 | 80.0 | 87.0 | 78.0 |
| BlitzNet300 (s4) | ResNet-50 | 78.2 | 86.8 | 85.1 | 78.3 | 70.4 | 47.5 | 85.4 | 85.0 | 86.2 | 59.0 | 81.8 | 77.9 | 86.9 | 86.1 | 85.4 | 78.6 | 54.9 | 81.9 | 81.1 | 87.7 | 78.2 |
| **BlitzNet300 + seg (s4)** | ResNet50 | **79.1** | 86.7 | 86.2 | 78.9 | 73.1 | 47.6 | 85.7 | 86.1 | 87.7 | 59.3 | 85.1 | 78.4 | 86.3 | 87.9 | 84.2 | 79.1 | 58.5 | 82.5 | 81.7 | 85.7 | 81.8 |
| SSD512* [16] | VGG-16 | 79.6 | 84.9 | 85.8 | 80.7 | 73.0 | 58.0 | 87.8 | 88.4 | 87.6 | 63.6 | 85.4 | 73.1 | 86.3 | 87.7 | 83.7 | 82.6 | 55.3 | 81.5 | 79.1 | 86.4 | 80.3 |
| **BlitzNet512 (s8)** | ResNet-50 | 80.7 | 87.7 | 85.4 | 83.6 | 73.3 | 58.5 | 86.6 | 87.9 | 88.5 | 63.7 | 87.3 | 77.6 | 87.3 | 88.1 | 86.2 | 81.3 | 57.1 | 84.9 | 79.8 | 87.9 | 81.5 |
| R-FCN[13] | ResNet-101 | 80.5 | 79.9 | 87.2 | 81.5 | 72.0 | 69.8 | 86.8 | 88.5 | 89.8 | 67.0 | 88.1 | 74.5 | 89.8 | 90.6 | 79.9 | 81.2 | 53.7 | 81.8 | 81.5 | 85.9 | 79.9 |
| Faster RCNN | ResNet-101 | 76.4 | 79.8 | 80.7 | 76.2 | 68.3 | 55.9 | 85.1 | 85.3 | 89.8 | 56.7 | 87.8 | 69.4 | 88.3 | 88.9 | 80.9 | 78.4 | 41.7 | 78.6 | 79.8 | 85.3 | 72.0 |
| YOLO [23] | YOLO net | 63.4 | - | - | - | - | - | - | - | - | - | - | - | - | - | - | - | - | - | - | - | - |
| **BlitzNet512 + seg (s8)** | ResNet50 | **81.5** | 87.0 | 87.6 | 83.5 | 75.7 | 59.1 | 87.6 | 88.0 | 88.8 | 64.1 | 88.4 | 80.9 | 87.5 | 88.5 | 86.9 | 81.5 | 60.6 | 86.5 | 79.3 | 87.5 | 81.7 |

Table 1: **Comparison of detection performance on Pascal VOC 2007 test set.** The models where trained on VOC07 trainval + VOC12 trainval. The models that have suffix "+ seg" where trained for segmentation jointly with data from VOC12 trainval and extra annotations provided by [8]. The values in columns correspond to average precision per class (%).

| network | backbone | mAP | aero | bike | bird | boat | bottle | bus | car | cat | chair | cow | table | dog | horse | mbike | persn | plant | sheep | sofa | train | tv |
|---|---|---|---|---|---|---|---|---|---|---|---|---|---|---|---|---|---|---|---|---|---|---|
| SSD300* [16] | VGG-16 | 75.8 | 88.1 | 82.9 | 74.4 | 61.9 | 47.6 | 82.7 | 78.8 | 91.5 | 58.1 | 80.0 | 64.1 | 89.4 | 85.7 | 85.5 | 82.6 | 50.2 | 79.8 | 73.6 | 86.6 | 72.1 |
| BlitzNet300 | ResNet50 | 75.4 | 87.4 | 82.1 | 74.5 | 61.6 | 45.9 | 81.5 | 78.3 | 91.4 | 58.2 | 80.3 | 64.9 | 89.1 | 83.5 | 85.7 | 81.5 | 50.5 | 79.9 | 74.7 | 84.8 | 71.1 |
| BlitzNet300 + COCO | ResNet50 | 80.2 | 91.0 | 86.5 | 80.0 | 70.1 | 54.7 | 84.4 | 84.1 | 92.5 | 65.1 | 83.5 | 69.2 | 91.2 | 88.1 | 88.5 | 85.7 | 55.8 | 85.4 | 79.3 | 89.8 | 78.2 |
| R-FCN[13] | ResNet-101 | 77.6 | 86.9 | 83.4 | 81.5 | 63.8 | 62.4 | 81.6 | 81.1 | 93.1 | 58.0 | 83.8 | 60.8 | 92.7 | 86.0 | 84.6 | 84.4 | 59.0 | 80.8 | 68.6 | 86.1 | 72.9 |
| Faster RCNN | ResNet-101 | 73.8 | 86.5 | 81.6 | 77.2 | 58.0 | 51.0 | 78.6 | 76.6 | 93.2 | 48.6 | 80.4 | 59.0 | 92.1 | 85.3 | 84.8 | 80.7 | 48.1 | 77.3 | 66.5 | 84.7 | 65.6 |
| YOLO [23] | YOLOnet | 57.9 | 77.0 | 67.2 | 57.7 | 38.3 | 22.7 | 68.3 | 55.9 | 81.4 | 36.2 | 60.8 | 48.5 | 77.2 | 72.3 | 71.3 | 63.5 | 28.9 | 52.2 | 54.8 | 73.9 | 50.8 |
| SSD512* [16] | VGG-16 | 78.5 | 90.0 | 85.3 | 77.7 | 64.3 | 58.5 | 85.1 | 84.3 | 92.6 | 61.3 | 83.4 | 65.1 | 89.9 | 88.5 | 88.2 | 85.5 | 54.4 | 82.4 | 70.7 | 87.1 | 75.6 |
| BlitzNet512 | ResNet50 | 79.0 | 89.9 | 85.2 | 80.4 | 67.2 | 53.6 | 82.9 | 83.6 | 93.8 | 62.5 | 84.0 | 65.8 | 91.6 | 86.6 | 87.6 | 84.6 | 56.8 | 84.7 | 73.9 | 88.0 | 75.7 |
| BlitzNet512 + COCO | ResNet50 | 83.8 | 93.1 | 89.4 | 84.7 | 75.5 | 65.0 | 86.6 | 87.4 | 94.5 | 69.9 | 88.8 | 71.7 | 92.5 | 91.6 | 91.1 | 88.9 | 61.2 | 90.4 | 79.2 | 91.8 | 83.0 |

Table 2: **Comparison of detection performance on Pascal VOC 2012 test set.** The models where trained on VOC07 trainval + VOC12 trainval. The BlitzNet models where trained for segmentation jointly with data from VOC12 trainval and extra annotations provided by [8]. Suffix '+ COCO' means that the model was pretrained on the COCO dataset. The reported values correspond to average precision per class (%). Detailed results of submissions are available on the VOC12 test server.

| network | seg | det | mIoU | mAP |
|---|---|---|---|---|
| BlitzNet300 | | ✓ | - | 78.9 |
| BlitzNet300 | ✓ | ✓ | **72.8** | **80.0** |
| BlitzNet300 | ✓ | | 72.4 | - |

Table 3: **The effect of joint learning on both tasks.** The networks where trained on VOC12 train-seg-aug, and tested on VOC12 val.

| network | seg | det | mIoU | mAP |
|---|---|---|---|---|
| BlitzNet300 | | ✓ | - | 83.0 |
| BlitzNet300 | ✓ | ✓ | **75.7** | **83.6** |
| BlitzNet300 | ✓ | | 72.4 | - |

Table 4: **The effect of extra data with bounding box annotations on segmentation performance.** The networks were trained on VOC12 trainval (aug) + VOC07 tainval. Detection performance is measured in average precision (%) and mean IoU is the metric for segmentation segmentation(%).

| network | seg | det | mIoU | mAP |
|---|---|---|---|---|
| BlitzNet512 | | ✓ | - | 33.2 |
| BlitzNet512 | ✓ | ✓ | **53.5** | **34.1** |
| BlitzNet512 | ✓ | | 48.3 | - |

Table 5: **The effect of joint training tested on COCO minival2014.** The networks were trained on COCO train.

| method | minival2014 | | | test-dev2015 | | |
|---|---|---|---|---|---|---|
| | int | 0.5 | 0.75 | int | 0.5 | 0.75 |
| BlitzNet300 | 29.7 | 49.4 | 31.2 | 29.8 | 49.7 | 31.1 |
| BlitzNet512 | 34.1 | 55.1 | 35.9 | 34.2 | 55.5 | 35.8 |

Table 6: **Detection performance of BlitzNet on the COCO dataset, with minival2014 and test-dev2015 splits** The networks were trained on COCO trainval dataset. Detection performance is measured in average precision (%) with different criteria, namely, minimum Jaccard overlap between annotated and predicted bounding box is 0.5, 0.75 or integrated from 0.5 to 0.95 % (column "int").

| network | backbone | mAP % | FPS | # proposals | input resolution |
|---|---|---|---|---|---|
| Faster-RCNN[25] | VGG-16 | 73.2 | 7 | - | $\sim 1000 \times 600$ |
| R-FCN[13] | ResNet-101 | 80.5 | 9 | - | $\sim 1000 \times 600$ |
| SSD300*[16] | VGG-16 | 77.1 | 46 | 8732 | $300 \times 300$ |
| SSD512*[16] | VGG-16 | 80.6 | 19 | 24564 | $512 \times 512$ |
| YOLO [23] | YOLO net | 63.4 | 46 | - | - |
| BlitzNet300 (s4) | ResNet-50 | 79.1 | 24 | 45390 | $300 \times 300$ |
| BlitzNet512 (s8) | ResNet-50 | 81.5 | 19.5 | 32766 | $512 \times 512$ |

Table 7: Comparison of inference time on PASCAL VOC 2007, when running on a Titan X (Maxwell) GPU.

| Block type | mAP | mIoU |
|---|---|---|
| Hourglass-style [19] | 78.7 | 75.6 |
| Refine-style [22] | 78.0 | **76.1** |
| ResSkip (no res) | 78.4 | 75.3 |
| ResSkip (ours) | **79.1** | 75.7 |

Table 8: **The effect of fusion block type on performance, measured on detection (VOC07-test) and segmentation (VOC12-val)** The networks were trained on VOC12-train (aug) + VOC07 tainval, see Sec. 4.1. Detection performance is measured in average precision (%) and mean IoU is the metric for segmentation segmentation(%).

detection results on COCO test-dev2015 in Table 6. Our results are also publicly available on the COCO evaluation test server.

### 4.6. Inference Speed Comparison

In Table 7 and Figure 4, we report speed comparison to other state-of-the-art detection pipelines. Our approach is the most accurate among the real time detectors working 24 frames per second (FPS) and in the setting close to real time (19 FPS), it provides the most accurate detections among the counterparts, while also providing semantic segmentation mask. Note that all methods are run using the same GPU (Titan X, Maxwell architecture).

### 4.7. Study of the Network Architecture

The BlitzNet pipeline simultaneously operates with several types of data. To demonstrate the effectiveness of the ResSkip block, we set up the following experiment: we leave the pipeline unchanged while only substituting this block with another one. We consider in particular fusion blocks that appear in the state-of-the-art approaches on semantic segmentation. [19] [22] [26]. Table 8 shows that our ResSkip block performs similar or better (on average) than all counterparts, which may be due to the fact that its design uses similar skip-connections as the Backbone network ResNet50, making the overall architecture more homogeneous.

Optimal parameters for the size of intermediate representations in segmentation stream (64) as well as the number of channels in the upscale-stream (512) where found by using a validation set. We did not conduct experiments by changing the number of layers in the upscale-stream as long as our architecture is designed to be symmetric with respect to the convolutions and the deconvolutions steps. Reducing the number of the steps will result in a smaller number of layers in the upscale stream, which may deteriorate the performance as noted in [16].

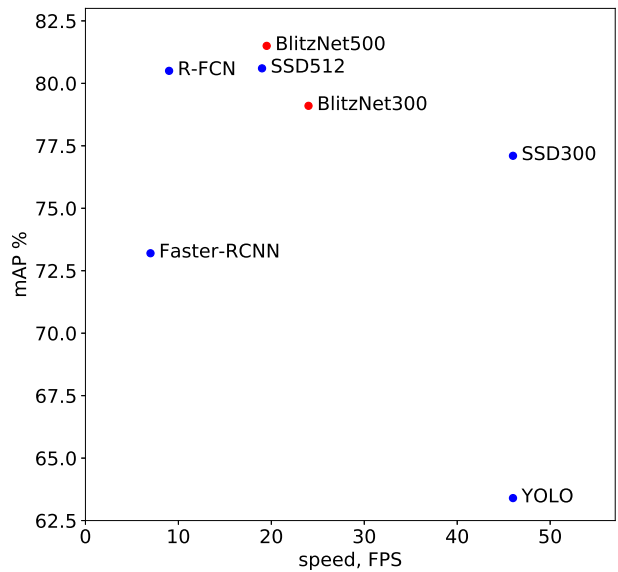

Fig. 4: **Speed comparison with other methods.** The detection accuracy of different methods measured in mAP is depicted on y-axis. x-coordinate is their speed, in FPS.

### 5. Conclusion

In this paper, we introduce a joint approach for object detection and semantic segmentation. By using a single fully-convolutional network to solve both problems at the same time, learning is facilitated by weight sharing between the two tasks, and inference is performed in real time. Moreover, we show that our pipeline is competitive in terms of accuracy, and that the two tasks benefit from each other.

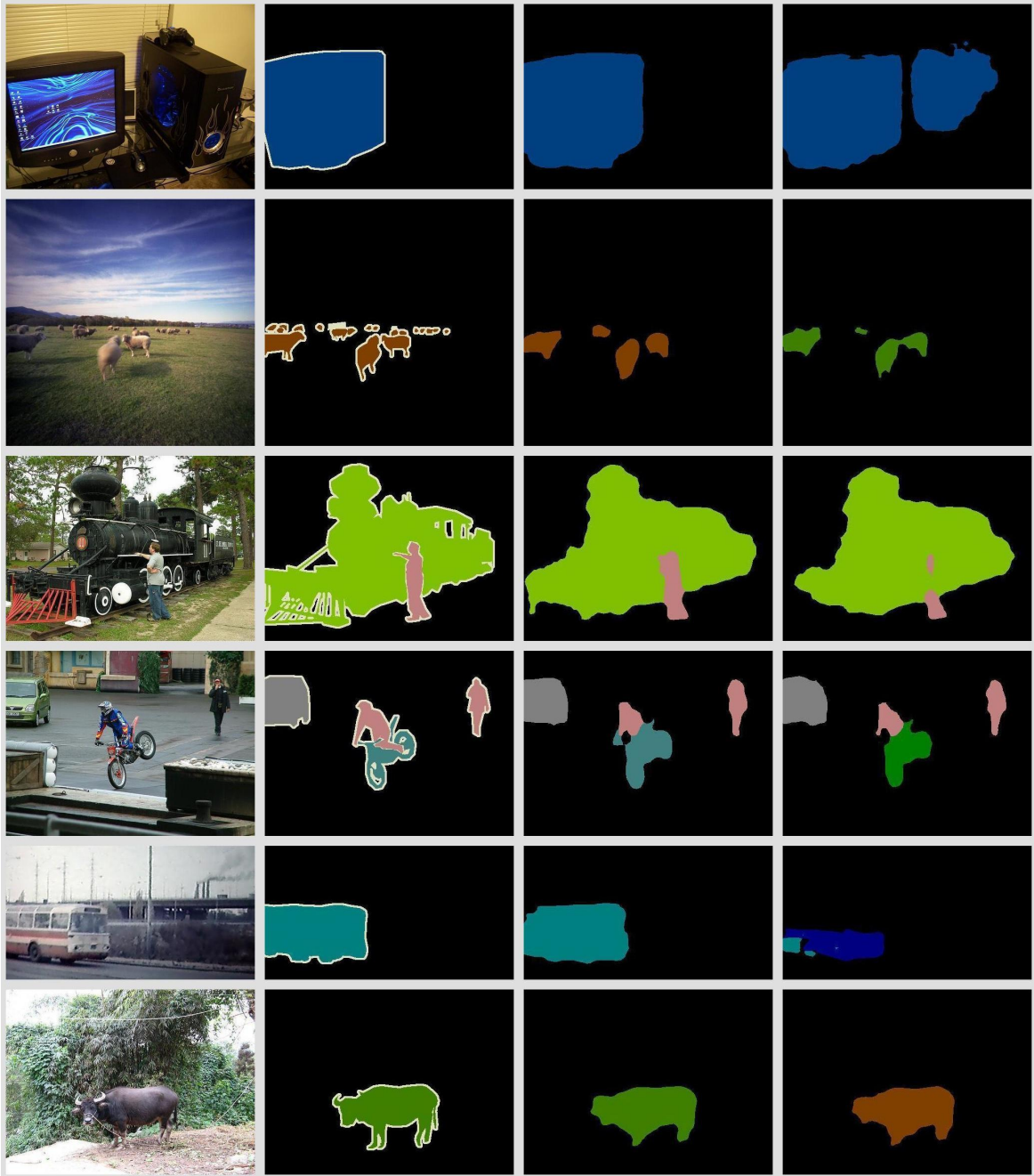

Fig. 5: **Effect of extra data annotated for detection on the quality of estimated segmentation masks.** The first column displays test images; the second column contains its segmentation ground truth masks. The third column corresponds to segmentations predicted by BlitzNet300 trained on VOC12 train-segmentation augmented with extra segmentation masks and VOC07. The last row is segmentation masks produced by the same architecture but trained without VOC07.

**Acknowledgements.** This work was supported by a grant from ANR (MACARON, ANR-14-CE23-0003-01) and by the ERC projects SOLARIS and ALLEGRO. We gratefully acknowledge the Intel gift and the support of NVIDIA Corporation with the donation of GPUs used for this research.

# Supplementary Material

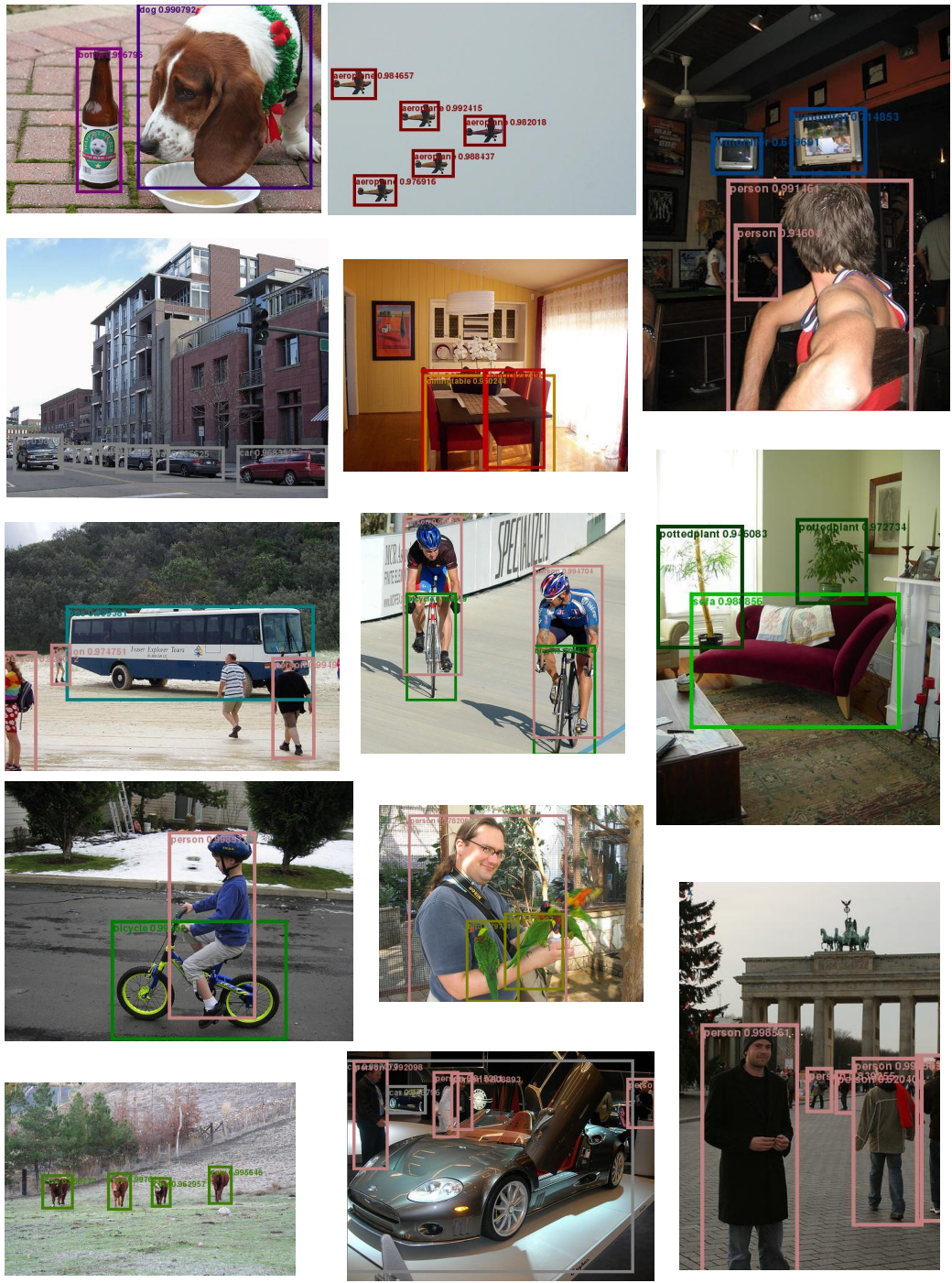

Fig. 6: **Qualitative results for the taks of object detection.** The results are obtained by the BlitzNet512 trained on VOC07 and VOC12 train-val augmented with extra segmentation masks.

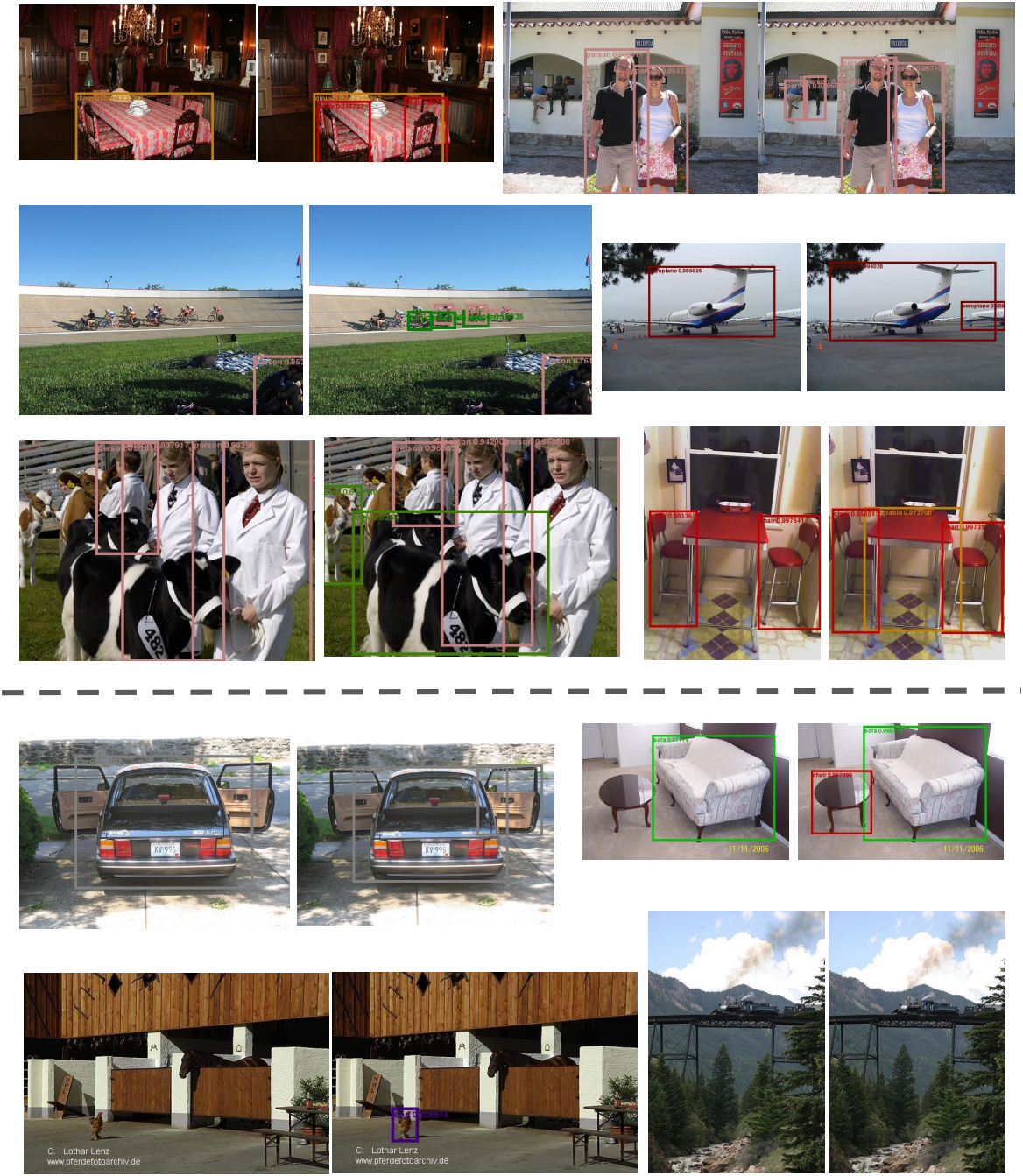

Fig. 7: **Improved and failure cases of detection by BlitzNet300 comparing to SSD300.** Each pair of images corresponds to the results of detection by SSD300 (left) and BlitzNet300 (right). The cases of improved detection are presened on the top part of the figure and the cases where both methods still fail are placed below the dashed line. It's clear that our pipeline provides more accurate detections in presence of small objects, complicated scenes and objects consisting of several parts with different appearance. The failure cases indicate that modern pipelines still struggle to handle ambiguous big objects (top left), intraclass variability (top right), misleading context (bottom right) and highly occluded objects (bottom left)